\documentclass[runningheads]{llncs}
\usepackage[T1]{fontenc}
\usepackage{graphicx}
\usepackage{cite}
\usepackage{amsmath,amssymb,amsfonts}
\usepackage{algorithmic}
\usepackage{graphicx}
\usepackage{textcomp}
\usepackage{xcolor}
\usepackage{subcaption}
\usepackage{float}
\newcommand{\highlight}[1]{\textbf{#1}}

\usepackage{hyperref}
\usepackage{CJKutf8}
\usepackage{multirow}
\usepackage{graphicx} 
\usepackage{pgfplots}
\usepackage{tikz}
\usepackage{booktabs}

\usepackage{flushend}
\begin{document}
\title{Deep Feature Embedding for Tabular Data}
\author{Yuqian Wu \and Hengyi Luo \and Raymond S. T. Lee\inst{*}\textsuperscript{,}\textsuperscript{\textdagger}, Member, IEEE}
\authorrunning{Wu et al.}
\institute{
Guangdong Provincial Key Laboratory of Interdisciplinary Research and Application for Data Science,\\ 
Beijing Normal University-Hong Kong Baptist University United International College,\\ 
Zhuhai 519000, China\\
\email{r130034042@mail.uic.edu.cn, q030026102@mail.uic.edu.cn}\\
\textsuperscript{\textdagger}\textit{corresponding author: raymondshtlee@uic.edu.cn}
}
\maketitle

\begin{abstract}
\textit{Tabular data learning} has extensive applications in deep learning but its existing embedding techniques are limited in numerical and categorical features such as the inability to capture complex relationships and engineering. This paper proposes a novel deep embedding framework with leverages lightweight deep neural networks to generate effective feature embeddings for tabular data in machine learning research. For numerical features, a two-step feature expansion and deep transformation technique is used to capture copious semantic information. For categorical features, a unique identification vector for each entity is referred by a compact lookup table with a parameterized deep embedding function to uniform the embedding size dimensions, and transformed into a embedding vector using deep neural network. Experiments are conducted on real-world datasets for performance evaluation. 
\end{abstract}
\section{Introduction}
\label{sec:introduction}
Tabular data remains the most common and valuable data for a wide spectrum of businesses to date, including retail, finance, recommendation, fraud detection, healthcare~\cite{xdeepfm,tracer,armnet} and etc.
\textit{Tabular data learning} refers to the predictive modeling of tabular data~\cite{tabnet,revisiting}, which is highly structured with each row corresponding to an instance and each column to a specific feature.
Real-world tabular data typically involves a fixed set of columns, with both numerical and categorical features.
One of the main characteristics of tabular data is that these features are heterogeneous.
Specifically, each numerical feature is a scalar value, quantitatively measuring a specific dimension, whereas each categorical feature is an ordinal identifier, uniquely denoting a specific \textit{entity}, e.g., words in NLP, products in recommendation~\cite{xdeepfm,afn}, and so on.
Due to the heterogeneity of features, and particularly, possible large cardinality of categorical features, e.g., millions or even billions of user or product IDs, traditional learning models such as SVM and GBDT can not effectively handle tabular data.

Notably, \textit{deep tabular learning} via deep neural networks (DNNs) is gaining momentum over the past few years~\cite{tabnet,netdnf,armnet,revisiting}.
The reasons for the preference of deep tabular learning over shallow counterparts are manifold:
besides a larger modeling capacity to handle heterogeneous features, DNNs obviate the need for laborious feature engineering~\cite{revisiting}, and the entire model can be optimized end-to-end together with other data modalities, e.g., texts, images as is common in healthcare analytics~\cite{tracer,armnet}.
In this light, there have been growing interests in devising DNNs for tabular learning, such as tree-like DNNs~\cite{survey}, attention-based DNNs~\cite{tabnet,revisiting} and interaction-based DNNs~\cite{xdeepfm,armnet}.
At the core of the success of deep tabular learning is \textit{feature embedding}, which determines how the input information is passed to the deep models.
However, the importance of feature embedding is often underappreciated, and in particular, feature embedding for deep tabular learning is less studied in the literature.

Succinctly, feature embedding refers to representing each input feature with a corresponding continuous vector~\cite{non,dhe,autodis}.
In deep tabular learning, the initial step typically involves standardizing the input by embedding each numerical and categorical feature into a $d$-dimensional vector for subsequent deep learning modeling~\cite{xdeepfm,armnet,revisiting}. 
The standard way to embed a categorical feature is to index an embedding vector from a lookup table using the feature value~\cite{dhe,nis,revisiting}.
While the major problem with embedding lookup is that the embedding table can be huge, which can possibly dominate the model size when the cardinality of the corresponding categorical feature is large~\cite{esapn,dhe,nis}.
Some recent works have proposed to reduce the embedding parameter size by limiting the cardinality via hashing and sharing lookup entries~\cite{hashemb}, but the resultant embeddings are not unique for each categorical entity due to the collision incurrence and such interdependence inevitably undermine embedding effectiveness. 
As for numerical feature embedding, the common practice is to linearly scale a corresponding embedding vector with the feature value~\cite{afn,armnet,revisiting}.
Despite the simplicity, the \textit{linearly-scaled embedding} may not be effective and could become an \textit{information bottleneck} of the subsequent modeling.
Several other attempts have also been made for embedding numerical features, e.g., by handcrafting embedding features~\cite{widedeep,youtube} or discretizing the feature into a categorical feature for embedding lookup~\cite{hashemb}, which however are either less effective or not widely applicable.
This paper aims to address existing embedding techniques for deep tabular learning via a unified \textit{deep embedding} framework for numerical and categorical features, which is illustrated in Figure~\ref{tab:categorical_summary} and~\ref{tab:numerical_summary} upon three principles: 1) \textit{effectiveness} measures the extent to which the embedding can carry feature information and be compatible with the subsequent deep modeling, 2) \textit{efficiency} describe whether the embedding is computation and parameter, and 3) \textit{generality} refers to the robustness and applicability of the embedding across models, applications, and domains.

To achieve these objectives, we propose deep embedding with lightweight DNNs dedicated to respective features.
For more effective numerical embeddings, we introduce a two-step \textit{feature expansion} and \textit{deep transformation} embedding technique.
Specifically, each numerical feature is firstly expanded into a $d$-dimensional feature vector by scaling and shifting the feature value with a learnable \textit{embedding sensitivity} and \textit{embedding bias} vector respective, to capture richer intra-feature semantic meanings of the input from different representation spaces. 
The \textit{embedding bias} is attached to a corresponding feature which encodes the static representations across inputs globally. Then, a DNN with residual connection and \textit{exp-centered} activation~\cite{nam} is further introduced to capture complex non-linear interactions among the expanded features. 
Typically, effectiveness is the major concern for numerical embedding, while the extra computation and parameters incurred by our deep embedding are negligible as compared with the subsequent deep modeling.
In contrast, for categorical embedding, the main concerns are the parameter efficiency and effectiveness, as embeddings of those minority entities are less frequently trained in the conventional embedding lookup approach~\cite{autodis}.
To achieve more parameter-efficient and effective categorical embedding, we also introduce a two-step embedding of \textit{entity identification} and \textit{deep transformation} for categorical features.
The first step only needs to index a unique and much smaller identification vector, likewise via a lookup table; and the second step is to construct the full feature embedding via deep transformation with a DNN.
With such \textit{deep factorization} of the embedding lookup table, a considerable amount of embedding parameters can be \textit{compressed} into the DNN, and through \textit{collaborative learning} among all the categorical entities via the shared DNN, the resultant embedding can thus be much more effective than the simple embedding lookup. 

To the best of our knowledge, \textit{deep embedding} is the first unified framework that provides DNN-enhanced embeddings for tabular data. The proposed deep embedding framework is easy to integrate with mainstream platforms as a seamless replacement for existing feature embedding modules in tabular data learning. Our key contributions are summarized as follows.
\begin{itemize}
    \item We propose an effective two-step feature expansion and deep transformation embedding technique for numerical features.
    
    \item We propose a parameter-efficient and effective \textit{deep factorization} embedding technique for categorical features.
    
    \item We create a unified embedding framework for deep tabular learning, streamlining inputs for end-to-end training without extensive feature engineering.
    \item We develop a unified embedding framework for deep tabular learning, which standardizes inputs for end-to-end training with deep models without feature engineering.
    
    \item We conduct extensive experiments on real-world datasets to verify the effectiveness and efficiency of deep embedding.
    
\end{itemize}
\section{Related Work}
\label{sec:related work}

\noindent
\highlight{Deep Tabular Learning.}
Tabular data~\cite{tabulardata,survey, treebased, neuralnets, transformerssurvey} is a data type used in many real-world applications, e.g., recommendation ~\cite{xdeepfm,afn} and healthcare~\cite{tracer,armnet}, to extract insights from tabular data for advanced analytics~\cite{tabtransformer,tabnet,armnet,revisiting}.
Tabular data learning is mostly \textit{shallow} models, such as GBDT, while they scale poorly to categorical features of high cardinality. There are many deep models have been proposed for tabular data in recent years, which can be broadly categorized into enhanced-DNN~\cite{netdnf}, tree-like~\cite{survey}, attention-based~\cite{tabnet,tabtransformer,armnet} and interaction-based~\cite{xdeepfm,non,afn,armnet} DNNs.
Enhanced DNNs propose enhancements to vanilla DNNs for effective tabular learning, such as \textit{disjunctive normal form} in Net-DNF~\cite{netdnf} and \textit{scaled exponential linear units} in SNNs~\cite{survey} to replace conventional linear feature extractors; Tree-like models pursued to copy the success of tree-based models by imitation, such as differentiable oblivious decision tree ensembles in NODE~\cite{treebased} and the integration of trivial neural networks as \textit{weak learners} into iterative gradient boosting learning in GrowNet~\cite{tabulardata};
Following the success of attention-based models in various domains, several new attention mechanisms are proposed for tabular data, e.g., iterative selective attention on features in TabNet~\cite{tabnet} and multi-head gated attention in ARM-Net~\cite{armnet}. 
Although various architectures and mechanisms have been proposed for deep tabular learning, their success is to a large extent, contingent on input feature embedding, which determines the amount of information to be passed to these models.

\vspace{1mm}
\noindent
\highlight{Numerical Feature Embedding.}
Numerical embedding is the core of deep tabular learning, which enhances the semantic meaning of scalar feature value of each numerical feature for subsequent learning. The existing numerical embedding techniques can be categorized into three major groups: handcrafted embedding~\cite{youtube,widedeep}, linearly-scaled embedding~\cite{afn,armnet}, and discretization~\cite{autodis}. 
Many works such as Wide \& Deep~\cite{widedeep} and SNNs~\cite{treebased} directly pass the original feature to DNNs. 
Then, works like YouTube deep ranking model~\cite{youtube} encode numerical features using several handcrafted functions. 
These works~\cite{afn,armnet} use linearly-scaled embedding, which are efficient and general. 
There are also recent works proposed to discretize each numerical feature into a categorical feature and then embed it via lookup for effective embedding. 
For example, AutoDis~\cite{autodis} used soft discretize on each feature and calculate a weighted average of a fixed set of embeddings attached to this feature.
These embedding via discretization often requires tremendous engineering efforts to decide the right techniques and hyperparameters for each step despite higher effectiveness.  
Our proposed deep numerical embedding, by contrast, is generic and effective, which can be used as an off-the-shelf module and trained end-to-end with deep models. 

\vspace{1mm}
\noindent
\highlight{Categorical Feature Embedding.}
Learned embedding is the core technique for passing categorical features to deep models~\cite{xdeepfm,afn,non,armnet,dhe}. 
Categorical feature embedding can represent many inputs, e.g., users~\cite{xdeepfm,afn}, items~\cite{armnet,dhe}, words, or any other entities by learning a corresponding unique continuous vector. 
Since uniqueness is required to distinguish different entities, the number of embedding entries corresponds to the possible number of entities can lead to a large lookup table~\cite{esapn,dhe,nis}. There are many recent works propose to reduce the embedding size, the number of embeddings by sharing lookup entries via hashing~\cite{hashemb,hybridhash,dhe}. 
For instance, HashEmb~\cite{hashemb} proposes to use multiple hash functions, each of which corresponds to a lookup table for learned weighted aggregation; 
Hybrid Hashing~\cite{hybridhash} uses embedding lookup for frequent entities and double hashing for less frequent ones for reducing the embedding size; 
and DHE~\cite{dhe} uses thousands of hash function to convert the categorical feature back into a continuous vector and use a DNN for deep transformation as the proposed deep embedding.
However, hashing inevitably leads to collisions resulting in less effective embedding; 
and hashing typically requires great engineering efforts and many hyperparameters to be tuned and non-negligible extra computation~\cite{neuralnets}. 
Our proposed deep embedding for categorical features circumvents these issues by following a more general two-step design, where the
first \textit{feature identification} step requires considerably fewer
parameters, and the second \textit{deep transformation} makes the final embedding more effective through collaborative learning effect.

\section{Problem Formulation}
\label{sec:formulation}

This section is to formulate the deep feature embedding problem for tabular data learning. The predictive modeling is to learn a mapping function $f: \mathbf{x} \rightarrow y$, where $\mathbf{x}$ is the feature vector, and $y$ is the prediction target. The feature vector $\mathbf{x}$ consists of $M$ numerical features and $N$ categorical features:
\begin{equation}\label{eq:features}
\begin{split}
\mathbf{x} = [x^{(n)}_1, x^{(n)}_2, \dots, x^{(n)}_M, x^{(c)}_1, x^{(c)}_2, \dots, x^{(c)}_N ]
\end{split}
\end{equation}
\noindent
where $x^{(n)}_i \in \mathbb{R}$ is the $i$-th numerical feature, and $x^{(c)}_j \in \mathbb{N}$ is the $j$-th categorical feature.
Each numerical feature $x^{(n)}_i$ is a real number, typically normalized into $[0, 1]$ or $\mathcal{N}(0, 1)$; and each categorical feature $x^{(c)}_j$ is a natural number that uniquely identifies an entity, e.g., a specific movie or product.

\vspace{1mm}
\noindent
\highlight{Feature Embedding.}
Feature embedding is to encode each feature value separately into a corresponding $d$-dimensional continuous vector space for subsequent modeling. It can be defined by realizing mapping functions for respective features:
\begin{equation}\label{eq:mapping}
\begin{split}
    f^{(n)}_i &: x^{(n)}_i \mapsto \mathbf{x}^{(n)}_i, \mathbf{x}^{(n)}_i \in \mathbb{R}^{d} \\
    f^{(c)}_j &: x^{(c)}_j \mapsto \mathbf{x}^{(c)}_j, \mathbf{x}^{(c)}_j \in \mathbb{R}^{d}
\end{split}
\end{equation}

\noindent
where $f^{(n)}_i$ and $f^{(c)}_j$ are the embedding functions for the $i$-th numerical feature and $j$-th categorical feature respectively.
After embedding, both numerical and categorical features are standardized, and the resultant matrix $\mathbf{X} = [\mathbf{x}^{(n)}_1, \dots, \mathbf{x}^{(n)}_M, \mathbf{x}^{(c)}_1, \dots, \mathbf{x}^{(c)}_N ]$ will then be forwarded to the subsequent neural networks for further modeling, e.g., capturing feature interactions~\cite{xdeepfm,afn,armnet}.

\section{Deep Feature Embedding}
\label{sec:embedding}

This section introduces mainstream feature embedding methods for numerical and categorical features, discussing their strengths and weaknesses. Then, we present deep feature embedding techniques, analyzing their effectiveness, efficiency, and generality for deep tabular data learning.

\subsection{Numerical Feature Embedding}
\label{sec:numerical_embedding}

Feature embedding for a numerical feature is to derive a deterministic function that maps a scalar feature value to a $d$-dimensional vector as formulated in Equation~\ref{eq:mapping}. 
Since numerical features in tabular data often carry distinct semantic meanings in different scales, the feature values are typically normalized before embedding, e.g., via min-max linear transformation~\cite{afn,armnet} or z-score standardization~\cite{survey}.

Most traditional learning models and many deep models~\cite{widedeep,youtube} directly use the normalized feature values as inputs, namely \textit{no embedding}.
While for more recent DNN-based models, the common practice is to embed each numerical feature~\cite{xdeepfm,non,armnet,autodis,revisiting} into continuous vector representations.
Table~\ref{tab:numerical_summary} summarizes the existing numerical feature embedding techniques. 
\begin{table}[t!]
    \small
    \centering
    \renewcommand{\arraystretch}{1.}
    \caption{Summary of numerical embedding techniques.}
    \label{tab:numerical_summary}
    
    \resizebox{1.0\columnwidth}{!}{
        \begin{tabular}{ c | c c | c c c c c}
    
    \toprule[1.5pt]
    Category & Dim & Params  &   Effectiveness & Efficiency-C &   Efficiency-P & Generality \\
    
    \midrule[0.5pt]
    No Embedding &   1   &   0  &   Low    &  High & High  &   High   \\
    Handcrafted Emb~\cite{youtube} &   d   &   0  &   Medium    &  High & High  &   Low   \\
    Linearly-Scaled Emb~\cite{afn} &   d   &   d  &   Medium    &  High & High  &   High   \\
    Discretization \& Emb~\cite{autodis} &   d   &   $v \cdot d$  &   Medium    &  Medium & Medium  &   Medium   \\
    Deep Embedding(ours) &   d   &   $2d + n_w$  &   High    &  Medium & Medium  &   High   \\
    \bottomrule[1.5pt]
        \end{tabular}
    }
    \vspace{-4mm}
\end{table}

\subsubsection{Deep Embedding for Numerical Feature}
\label{sec:numerical_deep}
\begin{figure}[t]
    \centering
    \includegraphics[width=0.8\textwidth]{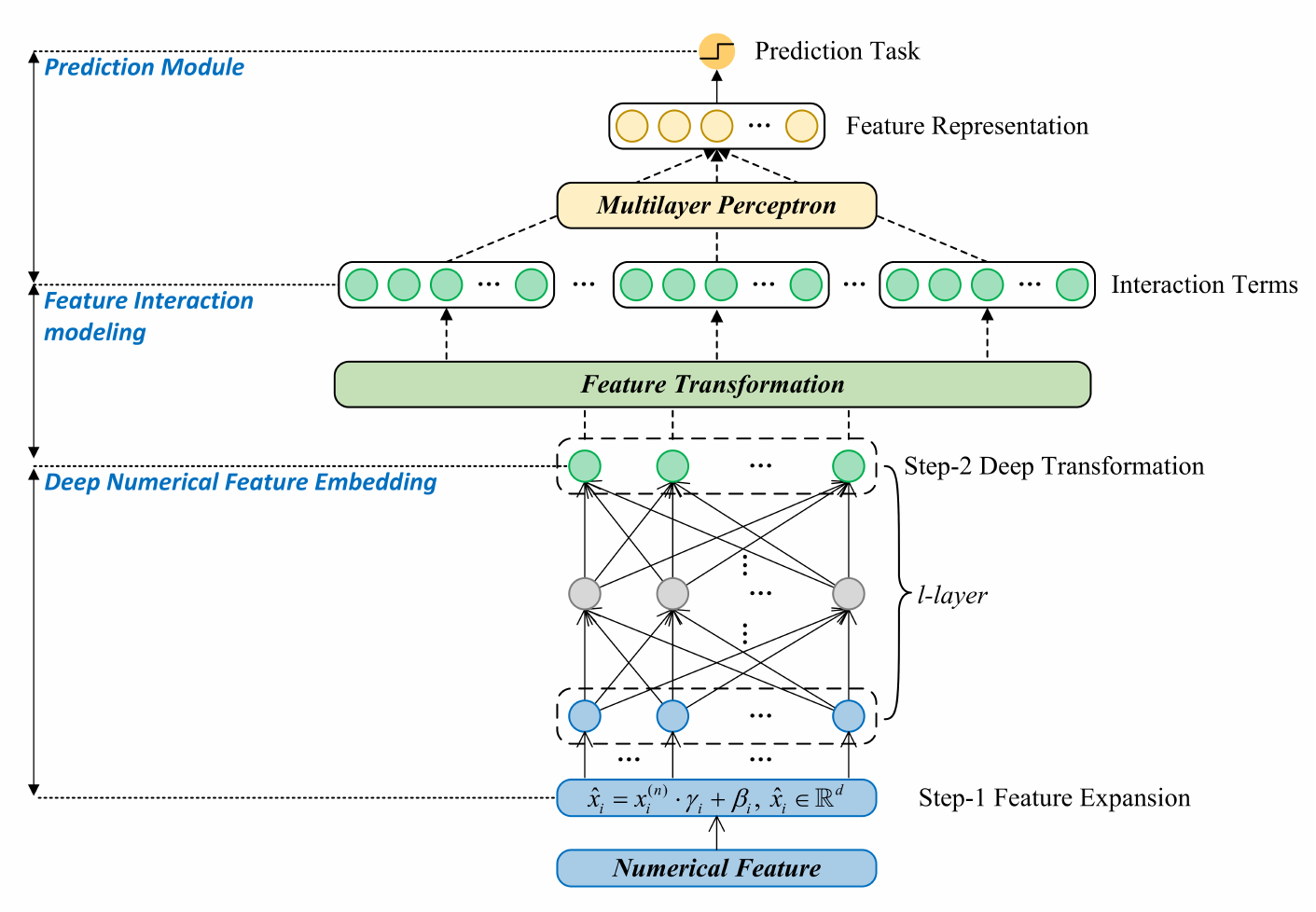}
    \caption{Deep numerical feature embedding framework.}
    \vspace{-5mm}
    \label{fig:deep-numerical-embedding}
\end{figure}

The proposed deep embedding framework is illustrated in Figure~\ref{fig:deep-numerical-embedding}. It consists of two steps. Note that all numerical features are required to be normalized prior embedding.

\vspace{1mm}
\noindent
\highlight{Step-1: Feature Expansion.}
Each input feature value $x^{(n)}_i$ is first expanded into a $d$-dimensional feature vector:
\begin{equation}\label{eq:numerical_expansion}
\begin{split}
    \mathbf{\hat{x}}_i = x^{(n)}_i \cdot \boldsymbol{\gamma}_i + \boldsymbol{\beta}_i, \mathbf{\hat{x}}_i \in \mathbb{R}^{d}
\end{split}
\end{equation}
\noindent
where $\boldsymbol{\gamma}_i, \boldsymbol{\beta}_i \in \mathbb{R}^{d}$ are learned \textit{embedding sensitivity} and \textit{embedding bias}.
In particular, each dimension of the output feature vector $\mathbf{\hat{x}}_i$ follows a different distribution: $\mathop{\mathbb{E}}[\mathbf{\hat{x}}_{ik}] = \boldsymbol{\gamma}_{ik} \cdot \mathop{\mathbb{E}}[x^{(n)}_i] + \boldsymbol{\beta}_{ik}$ and $\textrm{Var}[\mathbf{\hat{x}}_{ik}] = \boldsymbol{\gamma}_{ik}^2 \cdot \textrm{Var}[x^{(n)}_i]$, i.e., the mean and variance of the input feature $x^{(n)}_i$ are scaled and shifted by $\boldsymbol{\gamma}_{ik}$ and $\boldsymbol{\beta}_{ik}$, respectively.
Sine each pair of $\boldsymbol{\gamma}_{ik}$ and $\boldsymbol{\beta}_{ik}$ are learned independently, $\mathbf{\hat{x}}_i$ allows for representing a single feature value $x^{(n)}_i$ in $d$ feature dimensions, with different biases and sensitivities.
Further, in a way reminiscent of positional encoding, each embedding bias $\boldsymbol{\beta}_i$ is dedicated to a corresponding numerical feature, which learns the \textit{globally static} representations across inputs for this feature.
Experiment results showed that a simple effective \textit{embedding bias} enhancement to the \textit{linearly-scaled} approach can solely improve the fluency of numerical embedding.

\vspace{1mm}
\noindent
\highlight{Step-2: Deep Transformation.}
After expanding into a $d$-dimension feature vector $\mathbf{\hat{x}}_i$, we further propose to transform the embedding vector via a DNN:

\begin{equation}\label{eq:numerical_transformation}
\begin{split}
    \mathbf{x}^{(n)}_i = f^{(n)}_i( \mathbf{\hat{x}}_i; \mathbf{w}^{(n)}_i), \mathbf{x}^{(n)}_i \in \mathbb{R}^{d}
\end{split}
\end{equation}

\noindent
where the deep transformation $f^{(n)}_i: \mathbb{R}^{d} \mapsto \mathbb{R}^{d}$ is paramterized by $\mathbf{w}^{(n)}_i$ with $n_w$ parameters to be learned.
Specifically, $f^{(n)}_i$ is a $l$-layer feed-forward network (FFN) with residual connection~\cite{revisiting} and the \textit{exp-centered} (ExU) activation function~\cite{nam}.

The DNN $f^{(n)}_i$ is introduced to further capture complex non-linear interactions among the expanded features.
Thanks to the universal approximation capability of neural networks~\cite{nnapproximator,armnet,revisiting}, such deep transformation will further increase the modeling capacity and improve the embedding effectiveness.
We note that although additional parameters and computation are introduced, the embedding overhead is typically negligible as compared with the subsequent deep modeling.
Further, the whole deep embedding process can be vectorized, which can be readily implemented in any platform and accelerated by modern hardware such as GPUs.

\subsection{Categorical Feature Embedding}
\label{sec:categorical_embedding}

Categorical feature embedding as formulated in Equation~\ref{eq:mapping} is to determine a mapping function from an integer identifier to a $d$-dimension vector. Essentially, any entities can be embedded into continuous vector representations, e.g., words, products, or even a range of a numerical feature as discussed in Section~\ref{sec:numerical_deep}. 
Traditional learning models, such as SVM and GBDT, typically encode categorical features statically with \textit{no embedding}, e.g., via one-hot or binary encoding. 
In contrast, categorical embedding standardizes the input for the subsequent DNN-based modeling~\cite{xdeepfm,armnet,esapn,dhe,nis}, and the whole modeling pipeline can then be trained in an end-to-end manner.
This section will present mainstream categorical embedding techniques in Table~\ref{tab:categorical_summary} and present the proposed categorical embedding.
\begin{table}[b!]
    \small
    \centering
    \renewcommand{\arraystretch}{1.}
    \caption{Summary of categorical embedding techniques.}
    \label{tab:categorical_summary}
    
    \resizebox{1.0\columnwidth}{!}{
        \begin{tabular}{ c | c c | c c c c c}
    
    \toprule[1.5pt]
    Category & Dim & Params  &   Effectiveness & Efficiency-C    & Efficiency-P & Generality \\
    
    \midrule[0.5pt]
    Encoding - One-Hot &   v   &   0  &   Low    &  High & High  &   High   \\
    Encoding - Binary &   $\lceil \log(v) \rceil$   &   0  &   Low    &  High & High  &   High   \\
    Embedding Lookup~\cite{tabtransformer} &   d   &   $v \cdot d$  &   Medium    &  High & Low  &   High   \\
    Hashing \& Lookup~\cite{hashemb} &   d   &   $k \cdot \hat{v} \cdot d$  &   Medium    &  Medium & High  &   Low   \\
    Deep Embedding(ours) &   d   &   $v \cdot \hat{d} + \hat{n}_w$  &   High    &  Medium & High  &   High   \\
    \bottomrule[1.5pt]
        \end{tabular}
    }
\end{table}
\subsubsection{Deep Embedding for Categorical Feature}
\label{sec:categorical_deep}
\begin{figure}[t]
    \centering
    \includegraphics[width=0.8\textwidth]{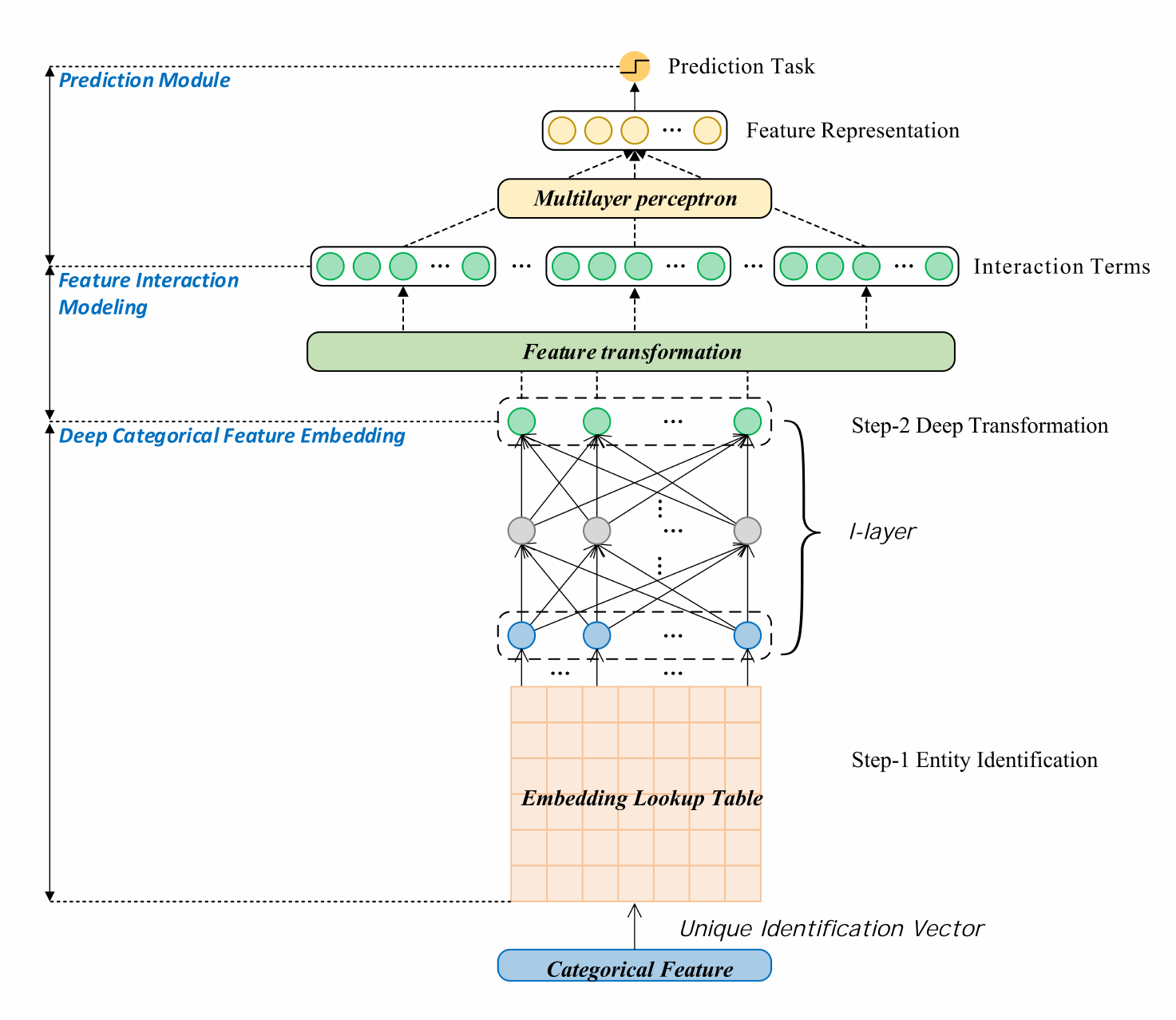}
    \caption{Deep categorical feature embedding framework.}
    \label{fig:deep-categorical-embedding}
\end{figure}

Despite the computational efficiency and generality, the main problems with embedding lookup are that firstly, the parameter size of the lookup table can be huge; and secondly, embeddings are learned separately and in a shallow manner, i.e., by simple indexing.
To make categorical embedding more parameter-efficient and effective, we propose deep categorical embedding with the following two steps and illustrate the deep embedding framework in Figure~\ref{fig:deep-categorical-embedding}.

\vspace{1mm}
\noindent
\highlight{Step-1: Entity Identification.}
Embedding each entity with an individual lookup entry is, to a large extent, inevitable for capturing the representations \textit{uniquely} and \textit{effectively}.
Therefore, we also adopt an embedding lookup table for each categorical feature.
While the lookup table here only needs to provide a unique identification vector for each entity.
As a consequence, the lookup table can be much smaller, specifically, $\mathbf{E}^{(c)}_j \in \mathbb{R}^{v_j \times \hat{d}}$, with $\hat{d} \ll d$.
The identification vector for $x^{(c)}_j$ is then $\mathbf{\hat{x}}_j = \mathbf{E}^{(c)}_j[x^{(c)}_j]$, $\mathbf{\hat{x}}_j \in \mathbb{R}^{\hat{d}}$.

\vspace{1mm}
\noindent
\highlight{Step-2: Deep Transformation.}
After obtaining the identification vector $\mathbf{\hat{x}}_j$, we propose to adopt a similar deep transformation as for numerical features: $\mathbf{x}^{(c)}_j = f^{(c)}_j( \mathbf{\hat{x}}_j; \mathbf{w}^{(c)}_j), \mathbf{x}^{(c)}_j \in \mathbb{R}^{d}$, where $f^{(c)}_j : \mathbb{R}^{\hat{d}} \mapsto \mathbb{R}^{d}$ is the deep embedding function paramterized by $\mathbf{w}^{(c)}_j$ with $\hat{n}_w$ learnable parameters.
The deep transformation is to firstly, restore the dimensionality back to a uniform embedding size $d$, and secondly, make the final embedding more effective with the strong modeling capacity of DNNs.

In essence, the deep transformation proposed here is a way to achieve \textit{model compression} and \textit{collaborative learning} via deep \textit{matrix factorization}.
Specifically, the conventional one-step embedding lookup is now factorized into two steps, where the first step only needs to learn a more parameter-efficient identification vector, and the second step is then to construct the final embedding with a shared DNN.
The highly redundant lookup table can thus be mostly compressed into the DNN, and the embedding parameter size shrinks from $v_j \cdot d$ to $v_j \cdot \hat{d} + \hat{n}_w$.
As the DNN is collaboratively learned and shared by all entities, the DNN embeddings can be more effective than simple individually-learned lookup ones.
Typically, the distribution of the entities is highly uneven, i.e., mostly following a power-law distribution, and therefore, the embeddings of infrequent entities are less indexed and trained as for embedding lookup.
While for the proposed two-step approach, the embeddings are mainly learned based on the shared DNN, and the knowledge can be transferred from other embeddings during training.
On the downside, the deep transformation incurs extra computation at runtime.
However, we note that the runtime computation can be considerably reduced by simply caching the embeddings of frequent entities, or, totally removed by reconstructing the entire lookup table once and for all via precomputation.

\section{Experiments}
\label{sec:experiment}
\begin{table}[b]
    \small
    \centering
    \renewcommand{\arraystretch}{1.}
    \caption {Dataset information.}
    \label{tab:dataset information}
    
    \resizebox{0.9\columnwidth}{!}{
        \begin{tabular}{ c | c c c | c }
    
    \toprule[1.5pt]
    Name &  Tuples &  Fields  &  Features & Task Type  \\
    
    \midrule[1pt]
    Frappe &   288609   &   10  &  5382  & Recommendation Systems   \\
    MovieLens &   2006859   &   3  & 90445  &   Recommendation Systems  \\
    Avazu &   40428967   &   22  &  1544250  &  Click-Through Rate Prediction  \\
    Criteo &   45302405   &   39  &  2086936  &   Click-Through Rate Prediction  \\
    Diabetes$_{130}$ &   101766   &   43  &   369   &  Healthcare \\
    \bottomrule[1.5pt]
        \end{tabular}
    }
    \vspace{-3mm}
\end{table}

\subsection{Experimental Setup}
\label{sec:experimental setup}

\vspace{1mm}
\noindent
\highlight{Dataset}
The proposed deep embedding framework used five real-world datasets on representative domains, namely app recommendation (Frappe), movie recommendation (MovieLens), click-through rate prediction (Avazu, Criteo), healthcare (Diabetes$_{130}$) for evaluation. These dataset properties are summarized in Table~\ref{tab:dataset information}.

\vspace{1mm}
\noindent
\highlight{Baseline Methods}
Experiments employed six distinct embedding methods~\cite{youtube, afn, autodis, tabtransformer, hashemb,tabulardata,neuralnets} across various backbone networks, including ARM-Net~\cite{armnet}, CIN~\cite{xdeepfm}, xDeepFM~\cite{xdeepfm}, DCN~\cite{dcn}, and two GBDT variants.
\label{sec: model architecture}
\begin{table*}[t]
    \small
    \centering
    \renewcommand{\arraystretch}{1.}
    \caption {Overall Prediction Performance with Various Embedding Methods in ARM-Net Model.}
    \label{tab:embedding method comparsion}
    
    \resizebox{1.0\textwidth}{!}{
        \begin{tabular}{ c c c c c c }
    
    \toprule[1.5pt]
    Embedding Method &  Frappe &  MovieLens  &  Avazu & Criteo & Diabetes$_{130}$ \\
    
    \midrule[1pt]
    Original Model Embedding &   0.9766   &   0.9548  &  0.7591  & 0.8086 & 0.6846  \\
    
    Handcrafted Embedding~\cite{youtube} &   0.9782   &   0.9481  & 0.7513  &   0.7857 & 0.6829  \\
    
    Linearly-Scaled Emb~\cite{afn} &   0.9789   &   0.9572  &  0.7651  &  \textbf{0.8092} & 0.6853  \\
    
    Discretization \& Emb~\cite{autodis} &   0.9763   &   0.9504  &  0.7524  &   0.7984 & 0.6765  \\
    
    Embedding Lookup\cite{tabtransformer} &   0.9792   &   0.9551  &   0.7656   &  0.8074 & 0.6846 \\
    
    Hashing \& Lookup~\cite{hashemb} &   0.9749   &   0.9547  &   0.7487   &  0.7888 & 0.6871 \\
    
    Deep Ensemble w/o GBDT~\cite{tabulardata} &  0.9746  & 0.9487 & 0.7393 & 0.7802 & 0.6764 \\
    
    Deep Ensemble w GBDT~\cite{tabulardata} &  0.9808  & 0.9567 & 0.7441 & 0.7902 & 0.6851 \\
    
    TabZilla Framework~\cite{neuralnets} &   0.9760   &   0.9560  &   0.7431   &  0.7832 & 0.6783 \\
    
    Deep Embedding(Ours) &   \textbf{0.9855}   &   \textbf{0.9597}  &   \textbf{0.7663}   &  0.8090 & \textbf{0.6895} \\
    \bottomrule[1.5pt]
        \end{tabular}
    }
    \vspace{-3mm}
\end{table*}

\vspace{1mm}
\noindent
\highlight{Evaluation Metrics}
We use the AUC (area under the ROC curve) metric to evaluate the effectiveness of our approach, where higher values indicate better performance, consistent with previous studies~\cite{armnet,xdeepfm,dcn}. 
Significant improvements at the 0.001 level were noted on the benchmark dataset. The dataset is divided into training set, validation set and test set in a ratio of 8:1:1. 
To mitigate randomness, we performed fifteen independent runs on the test set. Implementing early stopping during training, we give the average of the evaluation metrics.

\vspace{1mm}
\noindent
\highlight{Implementation details}
In the experiments, we trained our models with the Adam optimizer, using a learning rate of $0.1 \sim 1e-3$. For the Frappe and Diabetes$_{130}$ datasets, we employed a batch size of 1024, while for the remaining datasets, the batch size was set to 4096. For different numerical and categorical feature embedding methods, we follow the approaches provided in the original paper~\cite{youtube, afn, autodis,tabtransformer,hashemb,tabulardata,neuralnets} for handling categorical and numerical features to embed the data accordingly.

\vspace{1mm}
\noindent
\highlight{Hyperparameter Configuration}
For XGBoost and CatBoost, a Hyperopt library is used to optimize. For DCN, CIN, and xDeepFM, a search over interaction orders in 1 $\sim$ 8 is conducted , while for CIN and xDeepFM models, the number of features and neurons is explored within the range of $\{10, 25, 50, \dots, 800\}$. For ARM-Net model, the sparsity parameter $\alpha$ is searched in $1.0 \sim 2.0$. The final configuration is selected based on the set of hyperparameters corresponding to the smallest loss on the validation set. 

\vspace{1mm}
\noindent
\highlight{Experimental Environment}
The experiments are conducted in a server with Xeon(R) Gold 5218 CPU @ 2.30GHz (16 cores), NVIDIA A100 80G. Models are implemented in PyTorch 2.1.0 with CUDA 12.1.

\subsection{Overall Prediction Performance with Various Embedding Methods in ARM-Net Model}
\label{sec: comparing multiple embedding methods on the ARM-Net model.}
Experiments have used eight distinct embedding methods~\cite{youtube, afn, autodis,tabtransformer,hashemb,tabulardata,neuralnets} on ARM-Net model to  evaluate across five real-world datasets.
Based on the findings from Table~\ref{tab:embedding method comparsion}, our embedding framework has consistently demonstrated outstanding performance across the majority of datasets. 
Particularly notable is the performance on the Frappe dataset, where our method (0.9855) outperforms the second-best approach (0.9803) by 0.52\%, highlighting a significant margin of superiority. 
Despite Linearly-Scaled Embedding achieving the top result (0.8092) on the Criteo dataset, our proposed deep embedding approach (0.8090) still maintains competitiveness. 
This indicates that even when certain methods hold a slight advantage on specific datasets, our deep embedding approach remains effective in competing with them and even outperforms them on other datasets. 
We will delve deeper into the effectiveness and applicability of deep embedding methods in Section~\ref{sec: comparing deeplearning models and GBDT} and Section~\ref{sec:ablation study}.

\subsection{Comparing DL models and GBDT}
\label{sec: comparing deeplearning models and GBDT}
\noindent We evaluate the performance of GBDT and four deep learning models using our feature embedding framework on five public datasets using AUC as a metric. The results are shown in Table~\ref{tab:with and without}. 
We found that the embedding framework models improves performance over the original model on most datasets. They also perform better than GBDT on some datasets, such as Frappe, MovieLens, and Criteo. For instance, on the Frappe dataset, ARM-Net embedded in the framework achieves an AUC of 0.9855, which is 0.30\% higher than CatBoost's AUC of 0.9825, 1.48\% and 0.56\% respectively lower than XGBoost. 
Using the embedded framework, they outperform XGBoost by 0.15\% and 0.4\%. 
These findings show that our embedding framework enhances the information encoding and feature representation capabilities of deep learning models for tabular data, guiding the selection of appropriate models for different tasks and scenarios. 
It is worth noting that CatBoost achieved the best performance on the Diabetes$_{130}$ dataset with fewer features; we attribute this to the bias of neural networks towards overly smooth solutions and the impact of uninformative features on MLP-like neural networks more significantly~\cite{treebased}.

In summary, our study showcases the embedding framework's ability to enhance deep learning model performance in diverse tabular data scenarios. We found that these models outperform GBDT on select datasets and metrics, underscoring their competitiveness and potential. Our hope is that this work inspires further exploration and application of deep learning models in tabular data analysis.
\subsection{The Impact of Deep Transformation Layer Number on Model Performance}
\label{sec:layer number study}
\begin{figure}[t]
    \centering
    \begin{tikzpicture}
        \begin{axis}[
            width=5.5cm, 
            height=4.5cm,   
            line width=0.09pt, 
            xmin=1, 
            xmax=6,    
            ymin=0.9825, 
            ymax=0.9855,        
            xtick={1,2,3,4,5,6},  
            xticklabels={1,2,3,4,5,6}, 
            ytick={0.9825,0.9835,0.9845,0.9855},
            ymajorgrids=true,               
            grid style={dashed},       
            legend style={font=\fontsize{6}{12}\selectfont}, 
            enlarge x limits=0.05,  
            title={Frappe Dataset},
            scaled y ticks=false,
            y tick label style={/pgf/number format/.cd, fixed, precision=4}
        ] 
        \addplot[
                ybar,
                fill=blue
            ] coordinates {
                (1,0.9834)(2,0.9855)(3,0.9838)(4,0.9828)(5,0.9832)(6,0.9840)
            };
        \end{axis}
    \end{tikzpicture}
    \hspace{1cm}
    \begin{tikzpicture}
        \begin{axis}[
            width=5.5cm, 
            height=4.5cm,   
            line width=0.09pt,  
            xmin=1, 
            xmax=6,    
            ymin=0.6785, 
            ymax=0.6900,        
            xtick={1,2,3,4,5,6},  
            xticklabels={1,2,3,4,5,6}, 
            ytick={0.6785,0.6825,0.6860,0.6900}, 
            ymajorgrids=true,               
            grid style={dashed},       
            legend style={font=\fontsize{6}{12}\selectfont}, 
            enlarge x limits=0.05,  
            title={Diabetes$_{130}$ Dataset},
            scaled y ticks=false,
            y tick label style={/pgf/number format/.cd, fixed, precision=4}
        ]
        \addplot[
                ybar,
                fill=red
            ] coordinates {
                (1,0.6890)(2,0.6895)(3,0.6836)(4,0.6839)(5,0.6852)(6,0.6787)
            };
        \end{axis}
    \end{tikzpicture}
    \caption{Impact of Deep Transformation Layer size on ARM-Net Performance: AUC Evaluation on Frappe and Diabetes$_{130}$.}
    \label{fig:layer_size}
\end{figure}
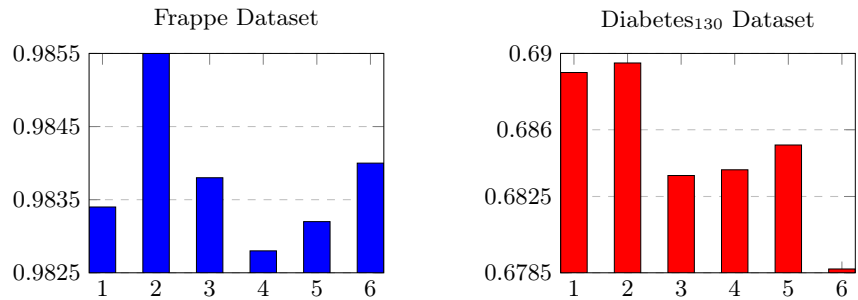

We studied the impact of changing the number of deep transformation layers (each with 500 neurons) on model performance, as shown in Figure~\ref{fig:layer_size}. 
In the Frappe dataset, two layers provide the best balance of performance, as more layers can lead to overfitting by capturing noise. 
In the Diabetes$_{130}$ dataset containing fewer features, increasing the number of transformation layers cannot significantly improve the representation ability, but will introduce too much noise. 
Therefore, we recommend designing model depth based on dataset characteristics for optimal performance.
\begin{table*}[t!]
    \small
    \centering
    \renewcommand{\arraystretch}{1.}
    \caption {The performance of each model varies across different datasets with (w) and without (w/o) Embedding Framework.
    }
    \label{tab:with and without}
    \resizebox{1.0\columnwidth}{!}{
        \begin{tabular}{|c|c|c|ccc|ccc|ccc|ccc|}
    \toprule[1.5pt]
    \multirow{2}{*}{\textbf{Dataset}}&\multirow{2}{*}
    {\textbf{XGBoost}} &\multirow{2}{*}
    {\textbf{CatBoost}} & \multicolumn{3}{c|}
    {\textbf{CIN}}&\multicolumn{3}{c|}{\textbf{xDeepFM}} &\multicolumn{3}{c|}{\textbf{ARM-Net}} &\multicolumn{3}{c|}{\textbf{DCN}}\\
    \cline{4-15}

     & & & w/o & w & Imprv. &w/o & w & Imprv. &w/o & w & Imprv. &w/o & w & Imprv.\\
    
    \midrule[1pt]
    Frappe & 0.9721 & 0.9825 & 0.9656 & {0.9766} & 1.10\% &  0.9773 & {0.9811} & 0.38\%  & 0.9766 & \textbf{0.9855} & 0.89\% & 0.9583 & 0.9796 & 2.13\%\\
    MovieLens & 0.9380 & 0.9548 & 0.9401 & {0.9516} & 1.15\%  &  0.9443 & {0.9491} & 0.48\%  & 0.9550 & \textbf{0.9597} & 0.47\% & 0.9401 & {0.9553} &  1.52\% \\
    Avazu & 0.7613 & 0.7634 & 0.6859 & {0.6915} & 0.56\%  &  0.6897 & {0.6953} & 0.56\%  & 0.7591 & \textbf{0.7663}& 0.72\% & 0.7460 & {0.7524} & 0.64\% \\
    Criteo & 0.7889 & 0.7867 & 0.7741 & {0.7904} & 1.63\%  & 0.7833& {0.7929} & 0.96\%  & 0.8086 & \textbf{0.8090} & 0.04\% &  0.7959 & {0.8083} & 1.24\% \\
    Diabetes$_{130}$ & 0.6753 & \textbf{0.7109} & 0.6776 & {0.6828} & 0.52\%  & 0.6683 & {0.6659} & -0.24\% &  0.6846 & {0.6895}& 0.49\% & 0.6765 & {0.6844} & 0.79 \% \\
    
    \bottomrule[1.5pt]
        \end{tabular}
}
\vspace{-5mm}
\end{table*}
\subsection{Ablation study}
\label{sec:ablation study}
This study conducted an ablation experiment to evaluate the performance of original deep learning models compared to models incorporating embedding frameworks. 
The experiment was conducted on five different datasets, namely Frappe, MovieLens, Avazu, Criteo, and Diabetes$_{130}$ with results listed in Table~\ref{tab:with and without}.
It demonstrate that the models with embedding frameworks consistently outperformed the original models across most datasets. 
Specifically, on the Frappe dataset, ARM-Net with our embedding framework achieved an AUC of 0.9855, significantly higher than the accuracy of 0.9766 obtained by original ARM-Net. 
On the MovieLens dataset, DCN and CIN with our framework attained AUCs of 0.9553 and 0.9516, respectively, surpassing the AUCs of original DCN (0.9401) and CIN (0.9401). Similarly, on the Criteo dataset, DCN with our embedding framework exhibited an AUC of 0.8083, outperforming the AUC of original DCN (0.7959). 
Lastly, on the Diabetes$_{130}$ dataset, DCN with the embedding framework demonstrated an AUC of 0.6844, outshining the AUC of original DCN (0.6765).

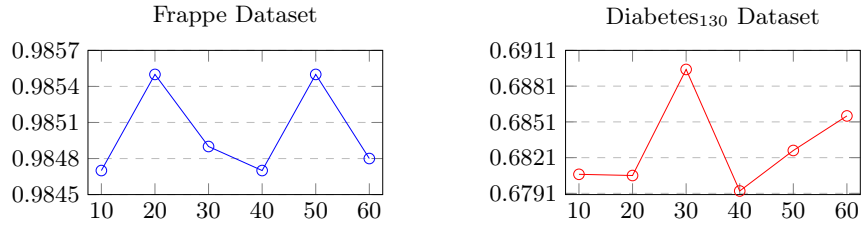
\begin{figure}[t]
    \centering
    \begin{tikzpicture}
        \begin{axis}[
            width=5.5cm, 
            height=3.5cm,   
            line width=0.09pt, 
            xmin=10, 
            xmax=60,    
            ymin=0.9845, 
            ymax=0.9857,        
            xtick={10,20,30,40,50,60},  
            xticklabels={10,20,30,40,50,60}, 
            ytick={0.9845,0.9848,0.9851,0.9854,0.9857},
            ymajorgrids=true,               
            grid style={dashed},       
            legend style={font=\fontsize{6}{12}\selectfont}, 
            enlarge x limits=0.05,  
            title={Frappe Dataset},
            scaled y ticks=false,
            y tick label style={/pgf/number format/.cd, fixed, precision=4}
        ] 
        \addplot[color=blue, mark=o]
            coordinates {
                (10,0.9847)(20,0.9855)(30,0.9849)(40,0.9847)(50,0.9855)(60,0.9848)
            };
        \end{axis}
    \end{tikzpicture}
    \hspace{1cm}
    \begin{tikzpicture}
        \begin{axis}[
            width=5.5cm, 
            height=3.5cm,   
            line width=0.09pt, 
            xmin=10, 
            xmax=60,    
            ymin=0.6790, 
            ymax=0.6911,        
            xtick={10,20,30,40,50,60},  
            xticklabels={10,20,30,40,50,60}, 
            ytick={0.6791,0.6821,0.6851,0.6881,0.6911},        
            ymajorgrids=true,               
            grid style={dashed},       
            legend style={font=\fontsize{6}{12}\selectfont}, 
            enlarge x limits=0.05,  
            title={Diabetes$_{130}$ Dataset},
            scaled y ticks=false,
            y tick label style={/pgf/number format/.cd, fixed, precision=4}
        ]
        \addplot[color=red, mark=o]
            coordinates {
                (10,0.6807)(20,0.6806)(30,0.6895)(40,0.6793)(50,0.6827)(60,0.6856)
            };
        \end{axis}
    \end{tikzpicture}
    \caption{Impact of Embedding Size on AUC Improvement for ARM-Net with Our Deep Embedding Framework on Frappe and Diabetes$_{130}$ datasets}
    \label{fig:impre--auc}
\end{figure}

Overall, these findings showed that models incorporating embedding frameworks generally outperform the original models on most datasets, with significant performance improvements (increases of over 1\%) observed in some cases, which indicates that embedding frameworks have a positive impact on information encoding and feature representation to offer an effective approach for deep learning models performance optimization. 
Notably, while models with embedding frameworks generally outperform the original models, there are instances where their performance is comparable or slightly inferior.
The effect of Embedding Size is also examined with ARM-Net model on Frappe dataset is illustrated in Figure~\ref{fig:impre--auc}. It showed that the benefits of incorporating embedding frameworks into deep learning models across multiple datasets. 

\section{Conclusion}
\label{sec:conclusion}
This paper proposed a novel deep embedding framework for tabular data to overcome limitations in handling numerical and categorical features. It used deep neural networks to enhance the effectiveness, efficiency, and generalization of feature embeddings with a two-step feature expansion and deep transformation technique for numerical features, and a parameter-efficient deep factorization embedding method for categorical features. Experimental results supported the efficacy of the proposed framework. Future research directions include exploring the embedding module optimization mechanisms methods for different functional transformations to each feature to expand the scope of diverse datasets and comparative analyses with other machine learning algorithms. These can further understand embedding's importance and potential in tabular deep learning.

\section{Acknowledgment}
\label{sec:acknowledgment}
The authors thank for Beijing Normal University-Hong Kong Baptist University United International College and the IRADS lab for the provision for computer facility for the conduct of this research.

\vspace{12pt}

\end{document}